\pdfoutput=1

\documentclass[11pt]{article}

\usepackage[final]{acl}

\usepackage{times}
\usepackage{latexsym}

\usepackage[T1]{fontenc}

\usepackage[utf8]{inputenc}

\usepackage{microtype}

\usepackage{inconsolata}

\usepackage{graphicx}
\usepackage{multirow}
\usepackage{amsmath}
\usepackage{xcolor}         
\usepackage{graphicx} 
\usepackage{booktabs} 
\usepackage{multirow} 
\usepackage{caption} 
\usepackage{amsfonts}       
\usepackage{bbding}

\title{Self-supervised Preference Optimization: \\Enhance Your Language Model with Preference Degree Awareness}

\author{
 \textbf{Jian Li\textsuperscript{1,}}\footnotemark[1],
 \textbf{Haojing Huang}\textsuperscript{1,}\footnotemark[1], 
 \textbf{Yujia Zhang\textsuperscript{1,}}\footnotemark[2],
 \textbf{Pengfei Xu\textsuperscript{1}},
 \textbf{Xi Chen\textsuperscript{2}}\footnotemark[2],
\\
 \textbf{Rui Song\textsuperscript{3}},
 \textbf{Lida Shi\textsuperscript{4}},
 \textbf{Jingwen Wang\textsuperscript{3}},
 \textbf{Hao Xu\textsuperscript{3,4}}
\\
\\
 \textsuperscript{1}AI Technology Center of OVB, Tencent, 
 \textsuperscript{2}Platform and Content Group, Tencent, \\
 \textsuperscript{3}College of Computer Science and Technology, Jilin University, \\
 \textsuperscript{4}School of Artificial Intelligence, Jilin University 
\\
 \small{
   \{loucasli,waterrhuang,yujiazhang,luciferxu,jasonxchen\}@tencent.com, \{songrui,xuhao\}@jlu.edu.cn,
 } \\
 \small{
 \{shild21,wjw22\}@mails.jlu.edu.cn
 }
}

\begin{document}
\maketitle

{\renewcommand{\thefootnote}{\fnsymbol{footnote}}%
\footnotetext[1]{These authors contributed equally to this work.}}

{\renewcommand{\thefootnote}{\fnsymbol{footnote}}%
\footnotetext[2]{Corresponding author}}

\begin{abstract}

Recently, there has been significant interest in replacing the reward model in Reinforcement Learning with Human Feedback (RLHF) methods for Large Language Models (LLMs), such as Direct Preference Optimization (DPO) and its variants. These approaches commonly use a binary cross-entropy mechanism on pairwise samples, i.e., minimizing and maximizing the loss based on preferred or dis-preferred responses, respectively. However, while this training strategy omits the reward model, it also overlooks the varying preference degrees within different responses. We hypothesize that this is a key factor hindering LLMs from sufficiently understanding human preferences. To address this problem, we propose a novel Self-supervised Preference Optimization (SPO) framework, which constructs a self-supervised preference degree loss combined with the alignment loss, thereby helping LLMs improve their ability to understand the degree of preference. Extensive experiments are conducted on two widely used datasets of different tasks. The results demonstrate that SPO can be seamlessly integrated with existing preference optimization methods and significantly boost their performance to achieve state-of-the-art performance. We also conduct detailed analyses to offer comprehensive insights into SPO, which verifies its effectiveness. The code is available at https://github.com/lijian16/SPO.

\end{abstract}

\section{Introduction}
\label{sec:intro}

The alignment of Large Language Models (LLMs) with human preferences is paramount, as it ensures that the outputs of LLMs are congruent with human values and ethical standards \cite{DBLP:conf/emnlp/BohmGMSDG19, DBLP:conf/emnlp/PerezKFWKC19, DBLP:journals/corr/abs-1909-08593}. Through meticulous tuning and ongoing learning of human preferences, LLMs can more accurately meet user needs while avoiding the generation of harmful or biased content \cite{DBLP:conf/nips/StiennonO0ZLVRA20, DBLP:journals/corr/abs-2402-11253}. Effective preference alignment not only enhances the applicability and safety of LLMs but also constitutes a critical step towards the responsible utilization of artificial intelligence.

To achieve human preference alignment of LLMs, a variety of methods have been developed. One prominent approach is Reinforcement Learning from Human Feedback (RLHF) \cite{DBLP:conf/nips/StiennonO0ZLVRA20, DBLP:journals/corr/abs-2204-05862}, such as Proximal Policy Optimization (PPO) \cite{DBLP:journals/corr/SchulmanWDRK17}, REINFORCE \cite{DBLP:journals/ml/Williams92} and their variants \cite{DBLP:conf/iclr/RamamurthyABHSB23}, which involve training reward models to optimize for objectives that are iteratively refined based on human feedback. However, these methods introduce increased complexity into the training process, involving training multiple models and sampling from the LLM in the loop of training \cite{DBLP:journals/corr/abs-2402-01306, DBLP:journals/corr/abs-2401-10020}. To streamline this process, recent works have proposed alternative solutions to reinforcement learning \cite{DBLP:journals/corr/abs-2309-06657, DBLP:journals/corr/abs-2305-10425, DBLP:journals/corr/abs-2402-01306,DBLP:journals/corr/abs-2310-12036}. DPO \cite{DBLP:conf/nips/RafailovSMMEF23} and its variants \cite{DBLP:journals/corr/abs-2309-16240, DBLP:journals/corr/abs-2402-09320, DBLP:journals/corr/abs-2402-01306, DBLP:conf/aistats/AzarGPMRVC24, DBLP:journals/corr/abs-2402-10571, meng2024simpo, yu2024direct} directly leverage pairwise responses to imbue the model with preference knowledge without a reward function. These methods achieve preference alignment by minimizing or maximizing the loss between each token in the language model's output and the tokens that are either preferred or not preferred. However, this training strategy overlooks a crucial aspect of a reward model: its ability to differentiate between varying degrees of human preferences in responses. We hypothesize that this is a key factor that prevents LLMs from fully understanding human preferences in those RLHF methods without a reward model.

\begin{figure*}[ht]
\centering
\includegraphics[width=1\textwidth]{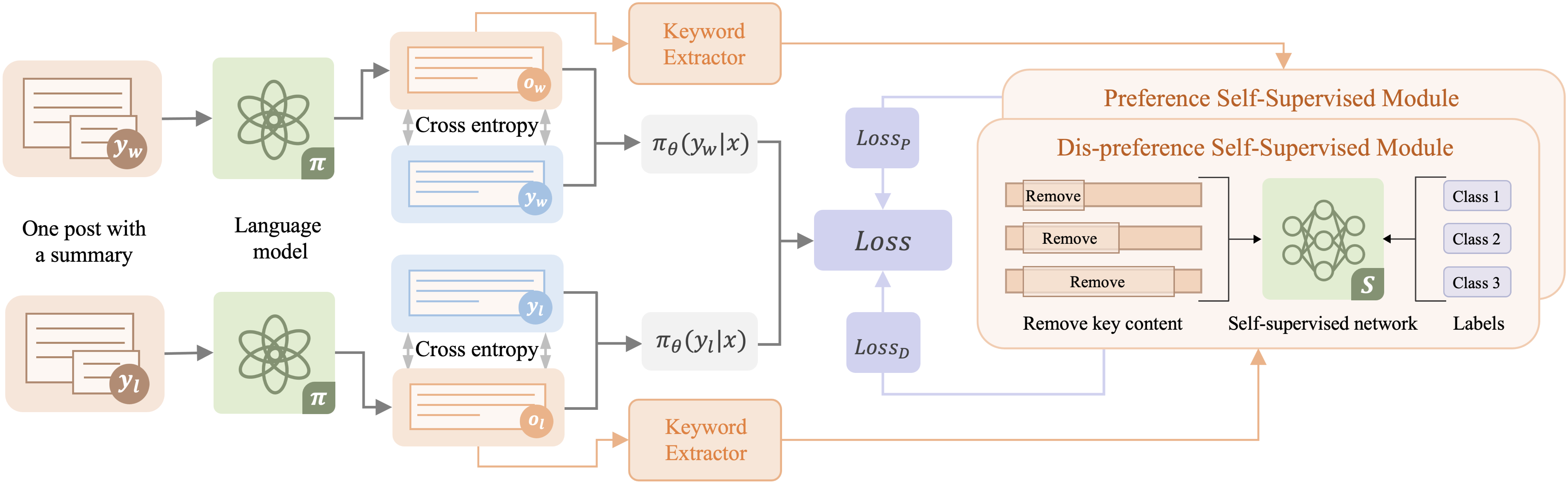}
\caption{The architecture of our proposed Self-supervised Preference Optimization (SPO) method involves employing an extractor to identify key content within the outputs of LLMs. Subsequently, self-supervised modules dedicated to preference and dis-preference randomly remove this content and undertake classification tasks. Ultimately, the loss derived from the classification is integrated with the alignment loss to jointly optimize the LLM.}
\label{Figure:Method}
\vspace{-0.5em}
\end{figure*}

To address this issue, we propose a novel Self-supervised Preference Optimization (SPO) scheme to help LLMs learn the degree of human preference and align LLMs with human preferences, simultaneously. The proposed method is illustrated in Figure \ref{Figure:Method}. Specifically, we design a novel auxiliary self-supervised task that selectively removes key content in LLM outputs to generate responses with varying degrees of preference. During the training process, we employ a keyword extractor \cite{automatic_keyword_extraction} on the outputs of LLMs to extract key content. By removing different amounts of the content, we construct responses with different degrees of preference. These responses are then fed into a self-supervised module for classification and the loss is integrated into the primary preference alignment loss (based on existing alignment methods) to jointly optimize LLMs. We observe that the key content within the LLMs' outputs is closely associated with preference information, as described in Section \ref{sec:analysis}. By gradually removing the content, we can effectively construct varying degrees of preferences. On the other hand, this method allows for the generation of multiple responses from a single output of LLMs, obviating the need for additional data collection and annotation efforts. We conduct comprehensive experiments on two widely used datasets of different tasks, i.e., Antropic HH \cite{DBLP:journals/corr/abs-2204-05862} and TL;DR summarization \cite{DBLP:conf/emnlp/VolskePSS17}. The results demonstrate that our proposed SPO can significantly enhance the performance of various existing alignment methods and achieve state-of-the-art results. Additionally, we conduct detailed analyses of multiple aspects and modules of our proposed SPO to provide comprehensive insights and verify its effectiveness.

The contributions of this work can be summarized as follows:

\begin{itemize}
    \item To our knowledge, we are the first to highlight a novel issue in direct human preference alignment methods: the binary training mechanism in these methods prevents LLMs from distinguishing varying degrees of preference, thereby limiting their performance.

    \item We innovatively propose a self-supervised preference optimization framework that can enhance human preference alignment performance without increasing any annotation or inference costs. This framework offers a novel approach to enhancing the performance of direct human preference alignment methods.

    \item Extensive experiments demonstrate that enhancing the ability of LLMs to distinguish degrees of preference can help improve performance across various tasks. SPO can be seamlessly integrated into existing alignment methods, significantly boosting them and achieving state-of-the-art results on two widely used datasets for different tasks.
    
\end{itemize}

\section{Method}
\label{sec:method}


In this section, we initially examine the pipeline of methods alternative to RLHF, with a primary focus on pairwise approaches that do not incorporate a reward model. Subsequently, we present the Self-supervised Preference Optimization (SPO), aimed at assisting LLMs in learning preference degrees at a fine-grained level.

\subsection{Preliminaries}
\label{subsec:preliminaries}


Methods alternative to RLHF generally avoid the process of learning rewards and consist of two stages: supervised fine-tuning (SFT) and preference optimization. These stages have seen extensive application in later research \cite{DBLP:journals/corr/abs-2305-10425, DBLP:journals/corr/abs-2402-01306}.

\textbf{SFT phase:} To tap into the capabilities of LLMs for particular tasks (e.g., summarization and dialogue), it is common practice to fine-tune a generically pre-trained LLM using supervised learning on a carefully curated dataset.



\textbf{Preference optimization phase:} The RLHF methods without a reward model typically start by gathering a pair of preferred $y_w$ and dispreferred $y_l$ responses for each prompt $x$. In the optimization process, these methods aim to make the LLM $\pi_{\theta}$ (initialized from the SFT model) produce a response that aligns more closely with $y_w$ and less so with $y_l$. To achieve this, the prompt $x$ is concatenated with both $y_w$ and $y_l$ separately as inputs, which are then fed into $\pi_{\theta}$ to generate predictions. These predictions are subsequently assessed by calculating the loss between them and $y_w$ as well as $y_l$. This loss is typically measured using the cross-entropy between each predicted token and its corresponding target token in the responses, as follows:
\begin{equation}
     \pi_{\theta}(y_\varepsilon|x)=-\frac{1}{K_{\varepsilon }} {\textstyle \sum_{i=1}^{K_{\varepsilon }}} \log P_{\theta}(y_{\varepsilon}^{(i)}|x,y_{\varepsilon}^{(<i)})
    \label{eq:pi_theta}
\end{equation}
\noindent whre $\varepsilon \in \{w,l\}$ and $K_{\varepsilon}$ denotes the number of tokens in $y_{\varepsilon}$, and $P_{\theta}(y_\varepsilon^{(i)}|x,y_\varepsilon^{<i})$ signifies the predicted probability of the $i^{th}$ target token in $y_\varepsilon$. The RLHF approach without a reward model primarily focuses on decreasing and increasing $\pi_\theta(y_w|x)$ and $\pi_\theta(y_l|x)$, respectively. Additionally, these methods employ a reference model $\pi_{ref}$ (e.g., a frozen SFT model) to mitigate deviation throughout the optimization process. Here, inputs are concurrently provided to $\pi_{ref}$ to calculate the corresponding loss $\pi_{ref}(y_\varepsilon|x)$. Based on these losses, such methods achieve their goal by the following loss function:
\begin{equation}
\begin{split}
    &  \mathcal{L}_{DPO}(\pi_{\theta},\pi_{ref}) = -\mathbb{E}_{(x,y_w,y_l)\sim \mathcal{D} } \\
    &  \Bigl[ \log\sigma(\beta \log \frac{\pi_{\theta(y_w|x)}}{\pi_{ref}(y_w|x)}) - \beta \log \frac{\pi_{\theta(y_l|x)}}{\pi_{ref}(y_l|x)})\Bigr] 
    \label{eq:dpo_loss}
\end{split}
\end{equation}
\noindent where $\sigma(\cdot)$ denotes a logistic function, such as the sigmoid function. The parameter $\beta$ regulates the extent of deviation from $\pi_{ref}$. While the specific operations employed by these methods vary, their core focus uniformly centers on $\pi_{\theta}(y_\varepsilon|x)$ \cite{DBLP:journals/corr/abs-2402-01306, DBLP:journals/corr/abs-2310-12036}. A more comprehensive discussion on alternative methods to RLHF is presented in Appendix~\ref{appendix:dpo_methods}.


\subsection{Self-supervised Preference Optimization}
\label{subsec:spo}

To grasp the degree of preference, we propose a straightforward Self-supervised Preference Optimization (SPO) method, which consists of preference extraction and self-supervised classification.


\subsubsection{Preference Extraction and Removing}
\label{subsec:per}

To facilitate the learning of preference degrees by LLMs, it is essential to provide them with a series of responses with different preference levels. To achieve this objective, existing methods commonly rely on generating multiple responses through one or more LLMs, subsequently employing manual efforts to annotate or rank these responses according to their preference levels \cite{DBLP:journals/corr/abs-2009-01325,DBLP:journals/corr/abs-2305-10425}. This process undeniably leads to an increase in both human labour and training expenses. To this end, we propose a novel and simple method for constructing preference data by extracting and removing key content from predictions of LLMs. From a semantic perspective, a sentence commonly contains key and additional content, where the former primarily dictates whether the sentence meets human preferences. Meanwhile, our experiments (described in Subsection \ref{subsec:reward_model_alignment}) reveal a close correlation between key content and preference information, indicating that adjusting the key content effectively modulates the degree of preference. Consequently, we try to extract the key content and gradually remove them to construct different responses. Specifically, during training, we decode all tokens predicted by LLMs into the corresponding text and then employ the Rapid Automatic Keyword Extraction (RAKE) \cite{automatic_keyword_extraction} to pinpoint the key content within the text. RAKE is an efficient, unsupervised method for the extraction of keywords from individual documents. It operates on a simple premise: keywords are typically content-bearing phrases that exclude common stop words and punctuation. The algorithm segments the document into candidate keywords $k$ and computes a score $S_k$ for each as follows:
\begin{equation}
    S_{k}= {\textstyle \sum_{w \in k}}(\frac{\mathrm{deg}(w)}{\mathrm{freq}(w)})
\end{equation}
\noindent where $\mathrm{deg}(w)$ is the degree of the word, representing its co-occurrence with other words within the candidate keyword, and $\mathrm{freq}(w)$ is the frequency of the word in the document. The candidate keywords with the highest scores are selected as the final keywords, providing a compact representation of its content suitable for various applications such as information retrieval systems and text analytics. 

Subsequently, we construct responses with different preferences by randomly removing a specified number of key contents from the predicted responses. Meanwhile, labels are assigned based on the number of removals: removing one item results in a label of 0, two items yield a label of 1, and so on. In this work, we introduce a self-supervised classification module with $N$ categories. Each category is associated with a specific level of content removal. During training, categories are randomly selected to dictate the extent of key content removal from the predictions. These modified predictions are then fed into the classification module for processing. To ensure a balanced representation of each category, we intentionally set an equal selection probability for every category. 




\begin{table*}[t]
\centering

\resizebox{1\linewidth}{!}{%
\begin{tabular}{@{}lccccccccc@{}}
\toprule

\multirow{2}{*}{Base model}  & \multicolumn{9}{c}{Antropic HH} \\ \cmidrule(l){2-10}
& DPO & +SPO & Incr. & IPO  &  +SPO &  Incr.  & KTO  & +SPO & Incr.  \\ \midrule

LLaMA-7B \cite{DBLP:journals/corr/abs-2302-13971} & 59.3\% & 62.1\% & {\color{red}+2.8\%} & 53.7\% & 56.4\% &  {\color{red}+2.7\%} & 60.7\% & 65.1\% & {\color{red}+4.4\%} \\
LLaMA-13B \cite{DBLP:journals/corr/abs-2302-13971} & 64.6\% & 67.8\% & {\color{red}+3.2\%} & 53.5\% & 57.2\% &  {\color{red}+3.7\%} & 64.2\% & 66.6\% & {\color{red}+2.4\%} \\
Mistral-7B \cite{jiang2023mistral} & 65.7\% & 67.9\% & {\color{red}+2.2\%} & 54.8\% & 57.7\% & {\color{red}+2.9\%} & 64.5\% & 68.1\% &  {\color{red}+3.6\%} \\
LLaMA-3-8B \cite{llama3modelcard} & 68.4\% & 71.1\% & {\color{red}+2.7\%} & 57.4\% & 61.2\% & {\color{red}+3.8\%} & 69.6\% & 72.8\% & {\color{red}+3.2\%} \\

\midrule

\multirow{2}{*}{Base model} & \multicolumn{9}{c}{TL;DR summarization} \\ \cmidrule(l){2-10}
& DPO & +SPO & Incr. & IPO  &  +SPO &  Incr.  & KTO  & +SPO & Incr.  \\ \midrule

LLaMA-7B \cite{DBLP:journals/corr/abs-2302-13971} & 81.0\% & 83.6\% & {\color{red}+2.6\%} & 50.4\% & 55.8\% &  {\color{red}+5.4\%} & 60.8\% & 65.4\% & {\color{red}+4.6\%} \\
LLaMA-13B \cite{DBLP:journals/corr/abs-2302-13971} & 82.8\% & 88.6\% & {\color{red}+5.8\%} & 55.2\% & 61.0\% &  {\color{red}+5.8\%} & 61.0\% & 65.8\% & {\color{red}+4.8\%} \\
Mistral-7B \cite{jiang2023mistral} & 86.6\% & 90.2\% & {\color{red}+3.6\%} & 56.5\% & 59.7\% & {\color{red}+3.2\%} & 57.8\% & 61.0\% & {\color{red}+3.2\%} \\
LLaMA-3-8B \cite{llama3modelcard} & 84.8\% & 88.0\% & {\color{red}+3.2\%} & 58.6\% & 61.2\% & {\color{red}+2.6\%} & 60.6\% & 64.3\% & {\color{red}+3.7\%}  \\
\bottomrule

\end{tabular}}
\caption{Comparative evaluation (\textit{win rate}) of advanced alignment methods and those with our SPO on Antropic HH (top) and TL;DR summarization (bottom) datasets.}
\label{table:main}
\end{table*}

\subsubsection{Self-Supervised Classification Modules}
\label{subsec:self_supervised_classification_modules}
To enhance LLMs' understanding of preference degrees, we introduce an innovative self-supervised preference classification module that improves preference awareness without incurring any additional labeling costs. Specifically, we first construct samples (using both preferred and dispreferred ground truth responses) with different preference degrees using our method in \ref{subsec:per}. The constructed samples are then fed into the self-supervised preference classification module to compute the preference classification loss, which is backpropagated together with the original DPO loss. The detailed architecture and operational processes of these modules are outlined below.

After extracting and removing key content from the predictions, we identify the corresponding tokens and hidden states of the remaining content. To help self-supervised classifier understand preference better, we propose to augment these hidden states $H=\{h_1,h_2,\dots,h_T\}$ from the last layer of LLMs with positional encoding before being fed into a Multilayer Perceptrons (MLP) \cite{lecun2015deep}, which can be defined as follows:
\begin{equation}
\begin{aligned}
    H_{pos}=H+P
\end{aligned}
\label{eq:h_pos}
\end{equation}
\noindent where $H_{pos}$ is the positionally encoded hidden states. Following \cite{DBLP:conf/naacl/DevlinCLT19}, the positional encoding $P$ can be computed as follows: 
\begin{equation}
\begin{aligned}
    P_{(pos, 2i)} &= \sin\left(\frac{pos}{10000^{2i/d}}\right) \\
    P_{(pos, 2i+1)} &= \cos\left(\frac{pos}{10000^{2i/d}}\right)
\end{aligned}
\label{eq:positional_encoding}
\end{equation}
\noindent where $pos$ denotes the position of a token (hidden state) in the sequence, $i$ for the dimension within the positional encoding, and $d$ as the size of the encoding vector. Subsequently, the hidden states $H_{pos}$ are fed into a projection layer following the design of \cite{chen2020simple, he2020momentum, DBLP:conf/nips/GrillSATRBDPGAP20} which outputs prediction probabilities $p$ for $N$ classes. The classification loss can be computed as follows:
\begin{equation}
    loss = -\sum_{i=1}^{N}y_i\log p_i
    \label{eq:cross_entropy}
\end{equation}
\noindent where $y$ represents the predefined self-supervised label based on one-hot encoding. Considering the implementation of two self-supervised modules, two classification losses are derived and then integrated with the main loss (e.g., $\mathcal{L}_{DPO}$) as follows:
\begin{equation}
    Loss = \mathcal{L}_{DPO} + \gamma*(loss_{pref} + loss_{dispref})
    \label{eq:main_loss}
\end{equation}
\noindent where $\gamma$ is a hyperparamter for scaling the classification losses $loss_{pref}$ and $loss_{dispref}$ from preference and dispreference modules, respectively.

\begin{figure*}[t]
\centering
\includegraphics[width=1\linewidth]{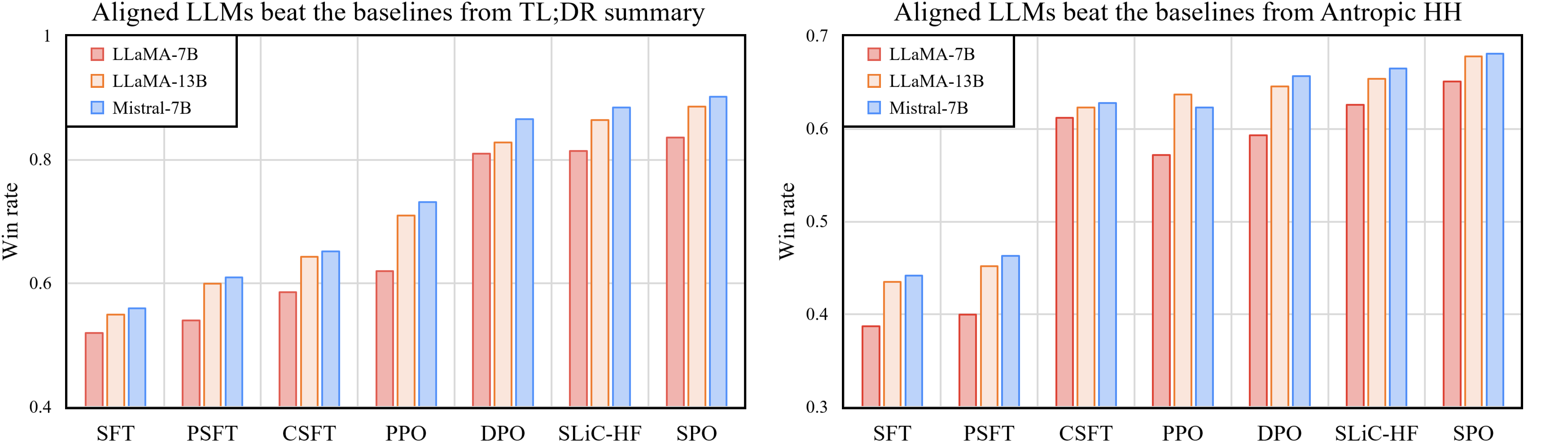}
\caption{Comparison of win rates with different state-of-the-art methods on TL;DR and Anthropic-HH datasets of three LLMs, i.e., LLaMA-7B, LLaMA-13B and Mistral-7B.}
\label{Figure:sota_results}
\end{figure*}

\section{Experiment}
\subsection{Settings}
\label{subsec:settings}

\textbf{Datasets}. In our experiments, two datasets designed for summarization and dialogue tasks are introduced, and LLMs are optimized using various alignment methods on the preference dataset $\mathcal{D} = \{ x^{(i)},y^{(i)}_w,y^{(i)}_l \}^N_{i=1}$. For the summarization task, the input $x$ denotes a forum post from Reddit\footnote{\url{https://reddit.com}}, and the LLMs are tasked with generating a succinct summary $y$ that captures the essence of the post. Following prior works \cite{DBLP:conf/nips/RafailovSMMEF23}, the Reddit TL;DR dataset \cite{DBLP:conf/emnlp/VolskePSS17} along with human preferences gathered by \citet{DBLP:journals/corr/abs-2009-01325} is employed. In the dialogue task, $x$ represents a human query, and LLMs need to produce an engaging and informative response $y$. The Antropic HH dataset \cite{DBLP:journals/corr/abs-2204-05862} is utilized, containing 170k dialogues between humans and automated assistants.

\textbf{Compared Methods}. To evaluate the efficacy of SPO in enhancing preference alignment, we extensively apply SPO to diverse existing methods (i.e., DPO~\cite{DBLP:conf/nips/RafailovSMMEF23}, IPO~\cite{DBLP:journals/corr/abs-2310-12036}, KTO~\cite{DBLP:journals/corr/abs-2402-01306}), as well as across different models, including Mistral-7B, LLaMA-7/13B and LLaMA3-8B. Furthermore, we also compare SPO with more methods which are recently published and representative of different frameworks for alignment. For example, methods based on SFT include Preferred SFT (PSFT) and Conditional SFT (CSFT) \cite{DBLP:conf/icml/KorbakSCBBPBP23}. Within the RLHF framework, PPO~\cite{DBLP:journals/corr/SchulmanWDRK17} is introduced. Additionally, SLiC-HF \cite{DBLP:journals/corr/abs-2305-10425} and SimPO \cite{meng2024simpo} are presented as alternative approaches to RLHF, functioning without a reward model. More details of these methods are described in Section \ref{sec:related_work}.

\textbf{Implemention}. In our experiments, all alignment methods are initialized from the SFT model. For the phase of SFT, a pre-trained LLM is fine-tuned over 2 epochs with a learning rate of 5e-5 and batch size of 64. For preference optimization, the SFT model is optimized for 1 epoch with a learning rate of 1e-5 and batch size of 32. For SPO, the classification number $N$ is set to 5 and the weight $\gamma$ is set to 0.1. The analysis of these hyperparameters is described in Section \ref{sec:analysis}. All experiments are conducted on 8 NVIDIA A100 GPUs. If it is not specifically mentioned, the settings of experiments that appear in this paper refer to this part.

\begin{figure*}[t]
\centering
\includegraphics[width=2\columnwidth]{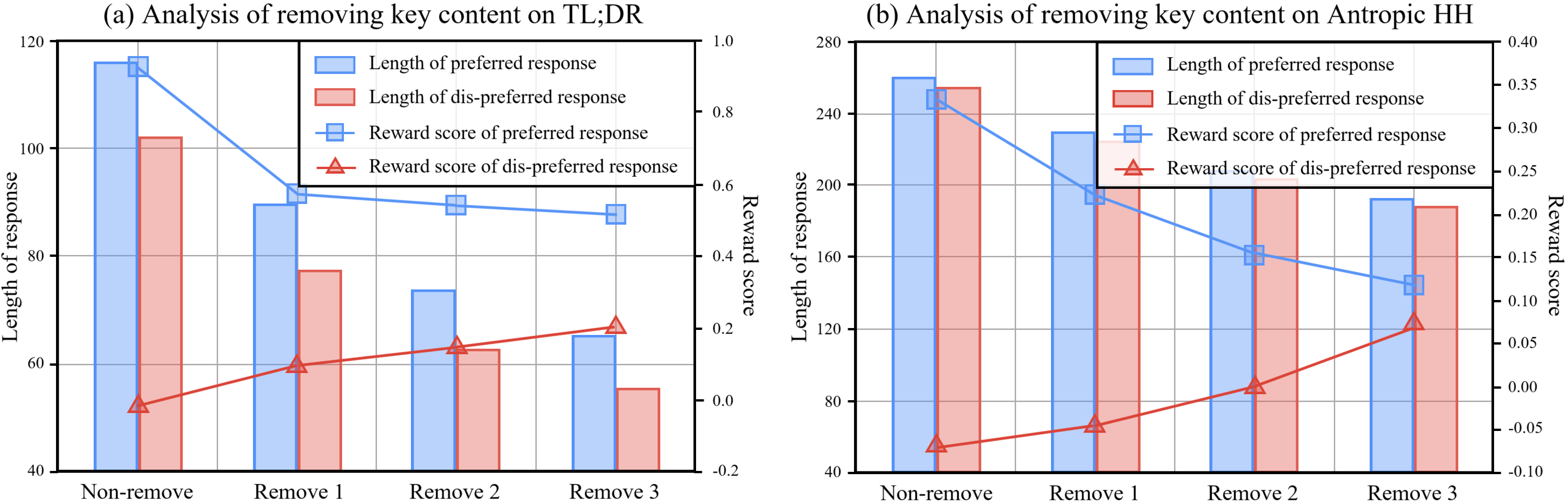}
\caption{Analysis of the relationship between key content and preferences on TL;DR and Antropic HH datasets.}
\label{Figure:mask_analysis}
\end{figure*}

\textbf{Metric}. Following \citet{DBLP:conf/nips/RafailovSMMEF23}, GPT-4 \cite{DBLP:journals/corr/abs-2303-08774} is employed to evaluate the generations of the aligned LLMs, i.e., comparing them with a baseline to determine which is more aligned with human preferences. The \textit{win rate} \footnote{The proportion of LLMs answers that GPT-4 prefers over the baseline preferences.} of these comparisons serve as the evaluation metric. For summarization, we use the reference summaries in the test set as the baseline, while the preferred responses within the test split serve as the baseline for dialogue. The detailed prompts of GPT-4 are shown in Appendix~\ref{appendix:prompts}.



\subsection{Main Result}
The results of the proposed SPO applied to existing alignment methods are shown in Table~\ref{table:main}. The results clearly demonstrate that SPO successfully improves the performance of all methods across both datasets. On TL;DR summarization dataset, we observe an average improvement of 4.04\% over the baseline methods. Notably, the LLaMA-7B model optimized with DPO+SPO surpasses the performance of the LLaMA-13B model optimized with DPO alone. Specifically, while the LLaMA-13B model optimized with DPO achieves a high win rate of 82.8\% on the TL;DR dataset, the proposed method further enhances this performance, achieving an impressive 5.8\% improvement. For the Anthropic HH dataset, our SPO also yields significant improvements. For instance, the LLaMA-7B model optimized with DPO+SPO shows a 2.8\% improvement over the DPO baseline, achieving a win rate of 62.1\%. Similarly, the LLaMA-13B model optimized with DPO+SPO achieves a win rate of 67.8\%, which is a 3.2\% improvement over the DPO baseline.
In addition to DPO, other alignment methods such as IPO and KTO also benefit from our SPO. Furthermore, as shown in Figure \ref{Figure:sota_results}, comparisons of SPO with other methods demonstrate its superiority in which SPO outperforms other methods and achieves state-of-the-art performance. Overall, SPO consistently enhances the performance of various alignment methods across different datasets and model sizes, demonstrating its effectiveness and robustness.


To further validate the effectiveness of our proposed method, we conducted additional experiments on two benchmark datasets commonly used in recent research on RLHF: Alpaca Eval 2.0 \cite{dubois_length-controlled_2024} and MT-Bench \cite{zheng_judging_2023}. Following the methodology of recent RLHF studies \cite{meng2024simpo,hong_orpo_2024,zhou_wpo_2024}, we trained an RLHF model on the Anthropic Helpful and Harmless (HH) dataset using Mistral-7B as the base model and evaluated it on Alpaca Eval 2.0 and MT-Bench. The results are summarized in Table\ref{table:Alpaca Eval/MT-Bench}.



\begin{table}[h]
    \small
    \centering
    \begin{tabular}{c|ccc@{}}
    \toprule
     \multirow{2}{*}{\textbf{Method}}  & \multicolumn{2}{c}{\textbf{Alpaca Eval 2.0}} & \textbf{MT-Bench} \\
     & LC Win Rate & Win Rate & Avg. Score \\
    \midrule
     DPO   & 5.20\%  & 2.91\%  & 2.98  \\
     DPO + SPO   & 5.65\% & 3.03\%  & 4.51 \\

    \bottomrule
    \end{tabular}
    \caption{Performance analysis on other datasets. To address length bias in evaluations, the Length-Controlled Win Rate (LC win Rate) metric is introduced.}
    \label{table:Alpaca Eval/MT-Bench}
\end{table}



The proposed SPO method significantly improved performance on both the Alpaca Eval 2.0 and MT-Bench benchmarks. Specifically, it increased the LC win rate by 0.45\% and the win rate by 0.12\% on Alpaca Eval 2.0, and boosted the average score by 0.53 on MT-Bench. These results validate the effectiveness of our method in enhancing performance on general tasks.

\section{Analysis}
\label{sec:analysis}

\subsection{Constructing Self-supervised Responses}
\label{subsec:constructing_self_supervised_data}


Our objective is to inject preference degrees into LLMs in a simple and efficient manner during the alignment process. To this end, the removal of specific content from predictions is proposed to effectively convey preference information. We hypothesize that different clauses or sub-words within the predictions contribute to preference degrees. By selectively removing certain elements, the preference levels can be altered accordingly. To validate this hypothesis, two strategies are explored: random removal and removal of key content. The results, presented in Table \ref{tab:removing_strategies}, demonstrate that both strategies yield performance improvements, suggesting that the model has successfully learned to represent preference levels. Notably, the key content extraction method outperforms random deletion in identifying content that significantly influences preference levels, thereby facilitating the construction of self-supervised responses with greater preference discrepancies. Of course, we also observe that such removal operations may compromise the semantic coherence of the responses. However, these responses are utilized solely as self-supervised classification signals rather than for direct preference alignment, with the objective of enabling LLMs to learn preference degrees. Meanwhile, experimental results on the HH and TL;DR datasets indicate that this approach does not introduce negative impacts.

\begin{table}[h]
    \small
    \centering
    \begin{tabular}{ccc@{}}
    \toprule
     Methods & Removal strategies & Win rate \\
    \midrule
     DPO   & -- & 81.0\%  \\
     DPO + SPO   & Random removal & 81.6\% \\
     DPO + SPO & Key content removal & 83.6\% \\

    \bottomrule
    \end{tabular}
    \caption{Analysis of different removal strategies for constructing self-supervised responses.}
    \label{tab:removing_strategies}
    \vspace{-1em}
\end{table}

\subsection{Analysis of Adjusting Key Content} 
\label{subsec:reward_model_alignment}

In this work, we extract key content from LLMs' predictions and then incrementally remove them to construct responses with varying preference degrees. To demonstrate its rationality, we first train two reward models initialized by LLaMA-7B on Antropic HH and TL;DR datasets, respectively, and further randomly sample 1,000 instances from each of these datasets. Following this, we extract their key content and sequentially remove 1-3 key elements from them to create four subsets with different preference intensities. The reward model is then employed to compute the average scores for these sets. The average score and length of each set are shown in Figure \ref{Figure:mask_analysis}. The experimental results indicate that as the number of key elements removed increases, the length of preference pairs gradually decreases. More importantly, the scores of preferred responses progressively decline, suggesting the preference information is being systematically eliminated. Conversely, the scores of dis-preferred responses exhibit an upward trend, as the dis-preferred information is being removed. These findings demonstrate the extracted key content accurately contains preference information and progressively removing these elements can construct responses with different preference intensities.


\begin{figure}[h]
\centering
\includegraphics[width=0.95\columnwidth]{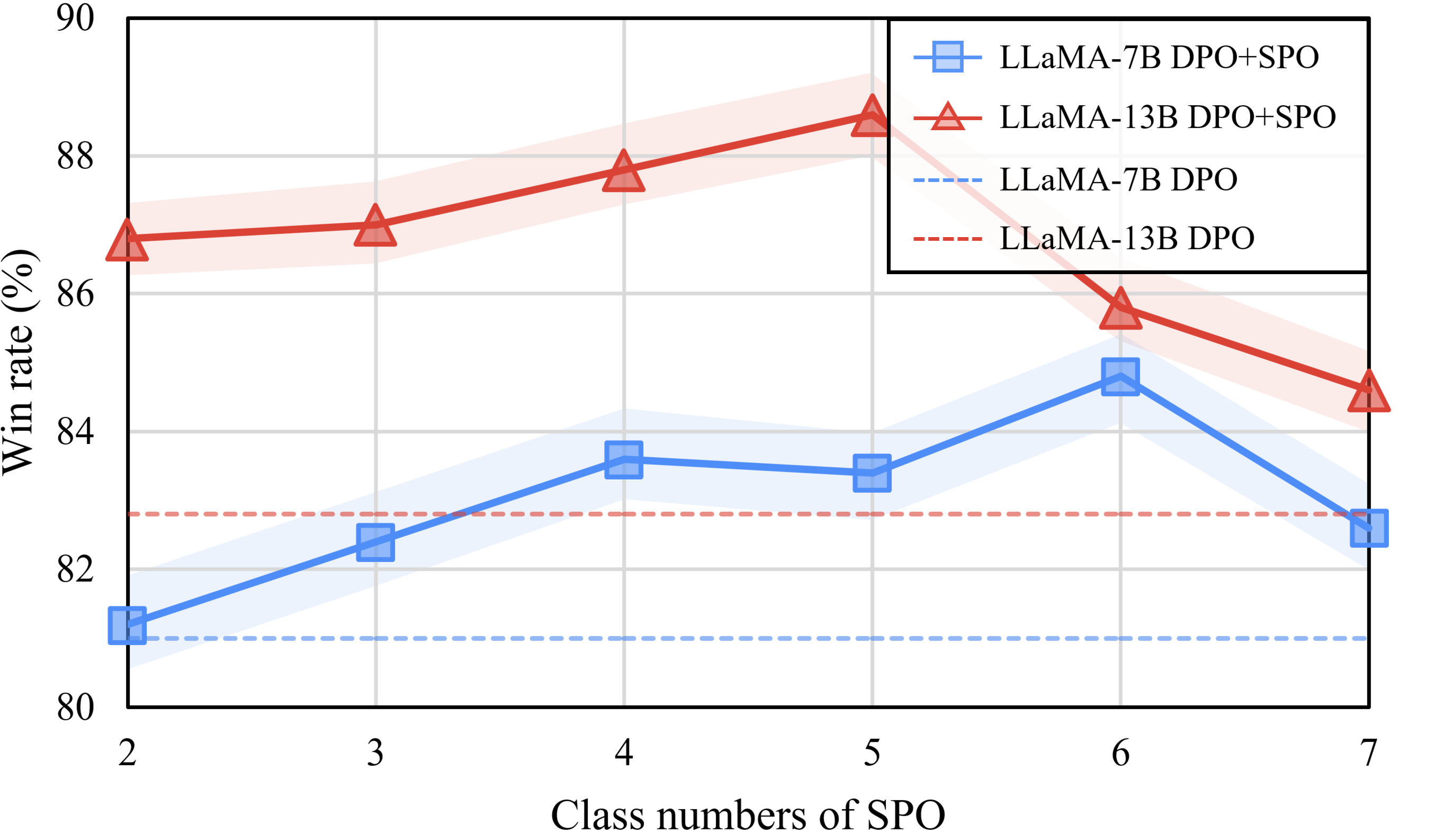}
\caption{The impact of self-supervised classification numbers on the performance. LLaMA-7B and 13B with DPO (+SPO) are trained on TL;DR dataset.}
\label{Figure:Ablation_classnum}
\vspace{-1em}
\end{figure}

\begin{table*}
    \centering
    \resizebox{1\linewidth}{!}{
    \begin{tabular}{@{}c|c|cccc@{}}
    \toprule
    \multirow{2}{*}{\textbf{Model}} & \multirow{2}{*}{\textbf{Method}} & \multirow{2}{*}{\textbf{Baseline}} & \textbf{+SPO} & \textbf{+SPO} & \textbf{+SPO} \\
    & & & \textbf{(Preference)} & \textbf{(Dis-preference)} & \textbf{(Both)} \\
    \midrule
    LLaMA-7B & DPO \cite{DBLP:conf/nips/RafailovSMMEF23} & 81.0\% & 82.8\% $^{\uparrow1.8}$ & 82.2\% $^{\uparrow1.2}$ & \textbf{83.6}\% $^{\uparrow2.6}$ \\
    LLaMA-7B & KTO \cite{DBLP:journals/corr/abs-2402-01306} & 60.8\% & 63.0\% $^{\uparrow2.2}$ & 64.2\% $^{\uparrow3.4}$ & \textbf{65.4}\% $^{\uparrow4.6}$ \\
    LLaMA-13B & DPO \cite{DBLP:conf/nips/RafailovSMMEF23} & 82.8\% & 87.2\% $^{\uparrow4.4}$ & 87.4\% $^{\uparrow4.6}$ & \textbf{88.6}\% $^{\uparrow5.8}$ \\
    \bottomrule
    \end{tabular}
    }
    \caption{Comprehensive analysis of the simultaneous implementation of dual self-supervised classification modules for preference and dis-preference.}
    \label{table:ablation_pre/dispre}
\end{table*}

\subsection{Analysis of Extracting Methods}
\label{subsec:analysis_extracting_method}

\begin{table}[h]
    \centering
    \resizebox{0.85\linewidth}{!}{
    \begin{tabular}{@{}ccc@{}}
    \toprule
    Methods & Methods for extracting & Win rate \\
    \midrule
     DPO   & -- & 81.0\%  \\
     DPO + SPO   & RAKE & 83.6\% \\
     DPO + SPO & YAKE & 81.6\% \\
     DPO + SPO & PositionRank & 80.8\% \\
    \bottomrule
    \end{tabular}
    }
    \caption{Analysis of various methods for extracting key content from the predictions from LLMs.}
    \label{tab:extracting_method}
    \vspace{-1em}
\end{table}

To identify an appropriate method for key content extraction, we investigate various extraction techniques (i.e., YAKE \cite{campos2020yake}, RAKE \cite{rose2010automatic} and PositionRANK \cite{florescu2017positionrank}) with DPO+SPO. The experimental results are summarized in Table \ref{tab:extracting_method} and examples of the extracted content are provided in Appendix \ref{app:case_extracting_method}. From the experimental results, we can see that SPO with RAKE and YAKE achieve 2.6\% and 0.6\% improvement in DPO, respectively, while SPO with PositionRank shows a 0.2\% decrease. From the examples, PositionRank extracts dispersed and incoherent key content, which likely makes it difficult for the classification module to learn preference degrees effectively, even resulting in a negative impact. YAKE, compared to PositionRank, extracts more continuous and complete key content, but it has issues with nested content. Although there is a 0.6\% improvement, it is relatively trivial. These experiments demonstrate the rationale for using RAKE.

\subsection{Self-supervised Classification Number}
\label{subsec:classification_num}

The classification number $N$ serves as a crucial hyperparameter within the self-supervised module. This study evaluates the impact of different $N$ on the performance of LLaMA-7B and 13B on the TL;DR dataset. As illustrated in Figure \ref{Figure:Ablation_classnum}, employing various values of $N$ consistently outperforms the baseline (i.e., LLaMA-7/13B with DPO), underscoring our method's efficacy. Specifically, the LLaMA-13B exhibits optimal performance with $N$ of 5, whereas further increasing the value of $N$ negatively affects performance. This trend suggests that a bigger $N$ complicates the classification task, thereby hindering effective learning. Similarly, the LLaMA-7B achieves its peak performance with $N$ of 6. These findings suggest choosing the number $N$ around 5 is a favourable option for alignment. 


\subsection{The Weight of Self-supervised Loss}
\label{subsec:weights_analysis}

\begin{figure}[h]
\centering
\includegraphics[width=1\columnwidth]{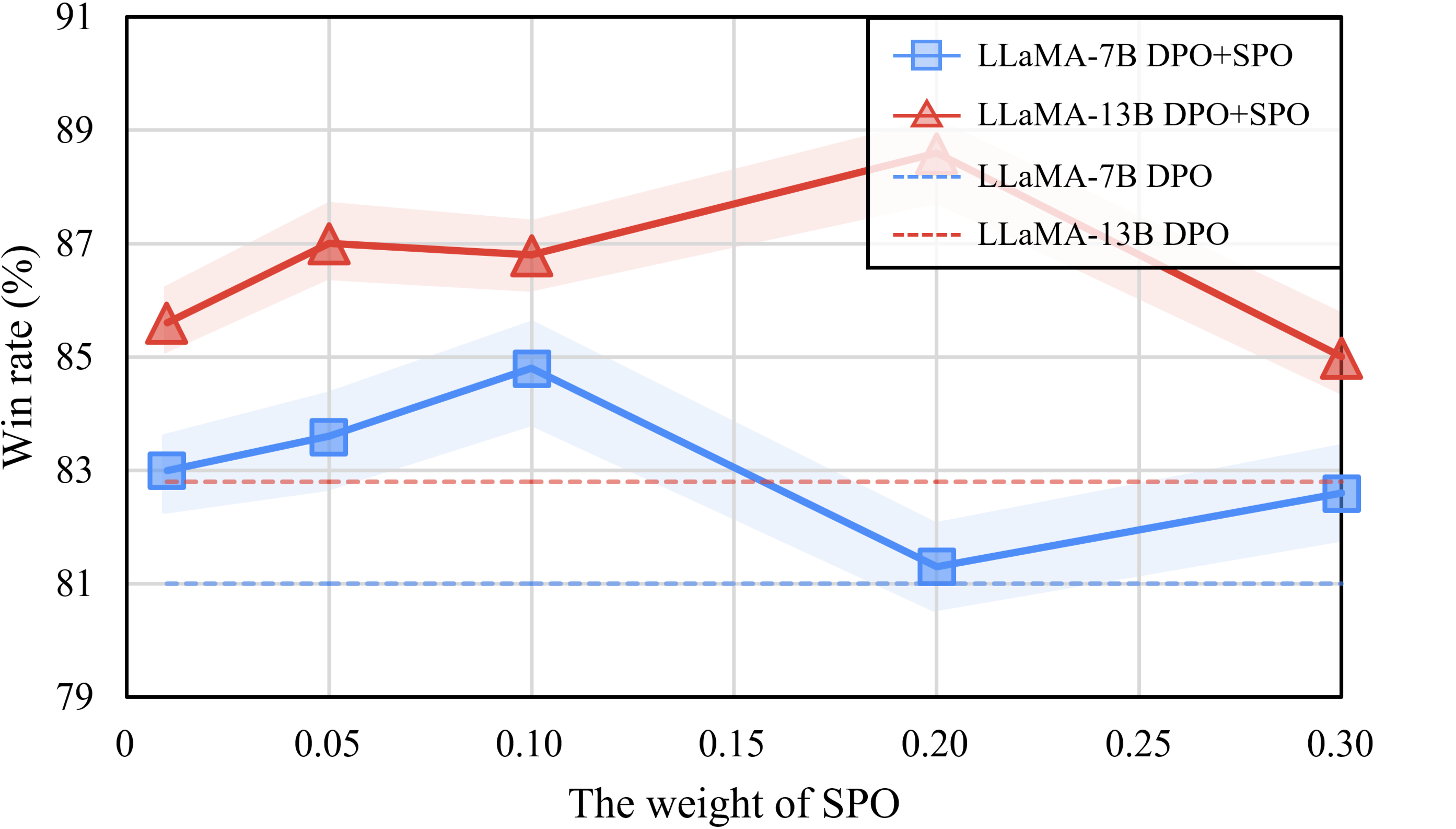}
\caption{The impact of the weight $\gamma$ on the performance. LLaMA-7B and 13B with DPO (+SPO) are trained on the TL;DR dataset.}
\label{Figure:Ablation_gamma}
\vspace{-1em}
\end{figure}

This study investigates the impact of weights $\gamma$ as defined in Equation~\ref{eq:main_loss} on the performance of LLaMA-7/13B using the TL;DR dataset. The findings, depicted in Figure \ref{Figure:Ablation_gamma}, reveal that excessively high weights detrimentally affect the performance of both models. Conversely, lower weights enhance the models' ability to assimilate information, thereby improving performance. Specifically, the LLaMA-7B demonstrates optimal performance with a weight of 0.1, whereas the LLaMA-13B achieves its best performance with a weight of 0.2. These results underscore the importance of carefully calibrating the weight of self-supervised loss to leverage its benefits without compromising the models' inherent performance capabilities.

\subsection{Analysis of Two Self-supervised Modules}
\label{subsec:preferred_modules}

In this work, we introduce two separate modules for preferred and dis-preferred predictions, respectively. To validate the combined efficacy of the modules, we additionally assess the impact of utilizing a single module for either preferred or dis-preferred prediction. As shown in Table~\ref{table:ablation_pre/dispre}, the results indicate that employing either the preference or dis-preference module independently enhances performance, however, simultaneous utilization of both modules yields a more substantial performance improvement. We consider that the concurrent application facilitates the sequential integration of preferred and dis-preferred intensity into LLMs without an excessive number of classes. Moreover, the merging of the two classification losses establishes a connection between preferred and dis-preferred information, enabling LLMs to learn coherent degree information from dis-preference to preference.

\subsection{Accuracy of Self-supervised Classification}

\begin{figure}
\centering
\includegraphics[width=1\columnwidth]{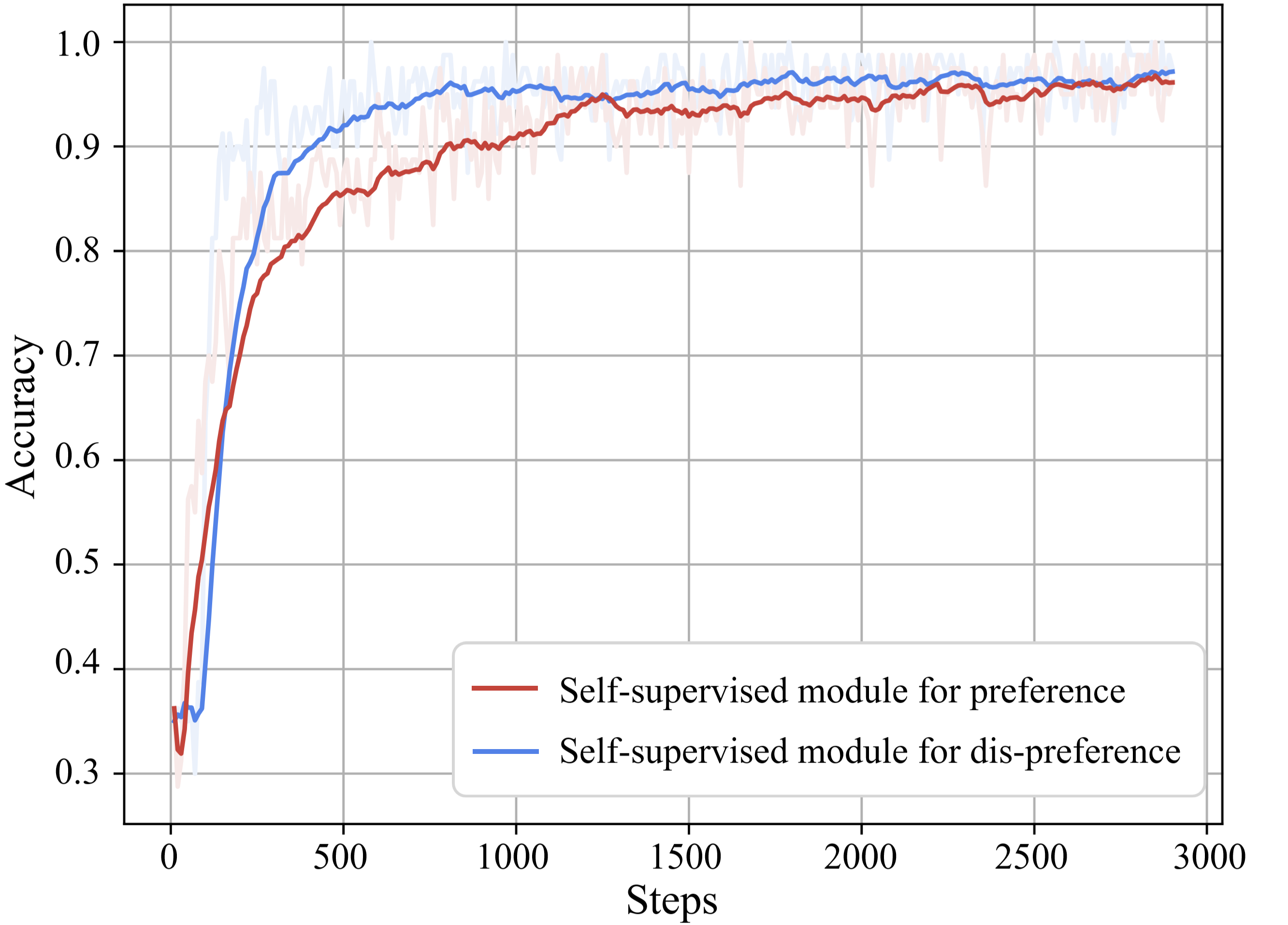}
\caption{Classification accuracy of self-supervised modules for preference and dis-preference, in which Mistral-7B with KTO+SPO is trained on TL;DR dataset.}
\label{Figure:self_supervised_accuracy}
\vspace{-1em}
\end{figure}

To assess whether the self-supervised modules function as intended, we evaluate their classification accuracy with KTO+SPO for Mistral-7B on the TL;DR dataset, as shown in Figure \ref{Figure:self_supervised_accuracy}. Within the first 1,000 steps, a significant upward trend in accuracy is observed, demonstrating that self-supervised modules can learn information related to preference intensity, thereby achieving precise classification. Subsequently, the accuracy of both modules stabilizes at over 90\%. This consistently high performance highlights the modules' ability to effectively capture and classify preference intensity, validating the usefulness of the self-supervised approach in preference alignment.



\section{Related Work}
\label{sec:related_work}

\subsection{Aligning LLMs with Human Preferences}

Preference alignment commonly begins with training a reward model on a preference dataset and further fine-tunes LLMs to maximize the identified reward by reinforcement learning, such as Proximal Policy Optimization (PPO) \cite{DBLP:journals/corr/SchulmanWDRK17}, REINFORCE \cite{DBLP:journals/ml/Williams92} and their variants \cite{DBLP:conf/iclr/RamamurthyABHSB23}. Although these methods effectively incorporate preference information into LLMs, they significantly complicate the training process in view of training multiple models and sampling from the LLM within the training loop \cite{DBLP:journals/corr/abs-2402-01306, DBLP:journals/corr/abs-2401-10020}. Following this, various methods have been proposed to streamline this process. For example, DPO \cite{DBLP:conf/nips/RafailovSMMEF23} bypasses the reward function to optimize LLMs by maximizing the difference between preferred and dispreferred responses. KTO \cite{DBLP:journals/corr/abs-2402-01306} streamlines the creation of preference pairs by optimizing the loss computation, eliminating the need for strict pairing between prompts and their preferred and dispreferred sequences. RSO \cite{DBLP:journals/corr/abs-2309-06657} suggests obtaining preference data from the estimated target optimal policy through rejection sampling in an offline manner. SimPO \cite{meng2024simpo} utilizes the average log probability of a sequence as an implicit reward and eliminates the need for a reference model, making it more compute and memory efficient.

While these methods show impressive performance, they overlook the degree of preference under a binary cross-entropy mechanism, which limits LLMs' ability to fully understand human preferences. In this work, we introduce a novel SPO framework to enhance LLMs' ability to learn human preference degrees in direct preference optimization methods, thereby improving their understanding capabilities of LLMs. 

\subsection{Self-Supervised Learning}
Self-Supervised Learning (SSL) has emerged as a powerful paradigm for leveraging unlabeled data to learn useful representations without explicit supervision \cite{DBLP:journals/tkde/LiuZHMWZT23,DBLP:conf/nips/LiangHCMS023,DBLP:conf/nips/YuanWSSLK23, zhang2022contrastive}. The foundational work of self-supervised learning can be traced back to the idea of using auxiliary tasks for which data itself provides supervision. \citet{DBLP:conf/nips/DosovitskiySRB14} introduces a novel approach where neural networks were trained to predict parts of the data given other parts, effectively learning representations without labelled data. This concept is further explored by \citet{DBLP:conf/eccv/NorooziF16}, who demonstrate that solving jigsaw puzzles as a pretext task could significantly improve feature learning. Following this line of thought, important self-supervised methods have emerged like mushrooms after rain and have had a profound impact on the field of deep learning research \cite{DBLP:journals/corr/abs-1807-03748, DBLP:conf/icml/ChenK0H20, DBLP:conf/iccv/ChenXH21, DBLP:conf/nips/GrillSATRBDPGAP20, khosla2020supervised, he2022masked}.


We integrate SSL into RLHF by leveraging self-supervised auxiliary tasks for the first time to enhance the comprehension abilities of LLMs.

\section{Conclusion}


In this work, we first identify a gap in alternative methods to RLHF, which overlooks the learning of preference degrees. To this end, we introduce a novel self-supervised preference optimization framework that integrates fine-grained human preference information into large language models (LLMs), thereby enhancing the understanding of human preferences. This approach does not require additional manual annotation and inference overhead. The proposed SPO can extract key content from the prediction of LLMs and selectively remove the content to construct responses with varying preference intensity. Subsequently, these responses are classified by the self-supervised modules and their losses are integrated with the alignment loss to jointly optimize LLMs. Extensive experiments and analyses fully demonstrate the effectiveness of our SPO.




\section*{Limitations} 

It would exist two limitations in this work. Firstly, the proposed SPO involves two hyperparameters $\gamma$ and $N$, for which the optimal settings vary across different methods and datasets, thereby undermining the convenience of SPO. In future work, we will explore adaptive hyperparameter tunning to tackle this issue. Furthermore, this work constructs responses with varying preference degrees by removing key content from predictions, which may compromise their semantic coherence. Although experimental results have demonstrated the effectiveness of our method, the potential impact on semantic integrity remains an area for further investigation. We will further explore construction method to minimize information distortion.

\section*{Ethics Statement}
While conducting our research on Self-supervised Preference Optimization (SPO), we are keenly aware of our ethical duties, including the prevention of misinformation and the protection of data privacy. The datasets in our experiments are all derived from publicly available information and we guarantee that we strictly adhere to the data usage policies outlined in the public datasets. In terms of self-supervised data construction, we ensure that no personal data is introduced, no manual labelling is involved, and we strictly adhere to privacy and data protection standards. In the experiments, we followed the evaluation methods in \cite{DBLP:conf/nips/RafailovSMMEF23, DBLP:journals/corr/abs-2402-01306}, using OpenAI APIs and strictly adhering to OpenAI's ethical and privacy protection guidelines. 

\bibliography{acl_latex}


\clearpage
\appendix


\section{Alternative Methods to RLHF}
\label{appendix:dpo_methods}

\subsection{Direct Preference Optimization}


The Direct Preference Optimization (DPO) method computes the losses associated with preferred (or dispreferred) responses by summing up the cross-entropy of each token in the preference answers alongside the matching token produced by LLMs, as described below:

\begin{equation}
    \begin{split}
       & \mathcal{L}_{DPO}(\pi_{\theta},\pi_{ref}) = -\mathbb{E}_{(x,y_w,y_l)\sim \mathcal{D} } \\ 
       & \left [ \log\sigma(\beta \log \frac{\pi_{\theta(y_w|x)}}{\pi_{ref}(y_w|x)} - \beta \log \frac{\pi_{\theta(y_l|x)}}{\pi_{ref}(y_l|x)})\right ] 
    \end{split}  
\end{equation}

\subsection{Sequence-Likelihood Calibration}


The Sequence-Likelihood Calibration (SLiC) employs a margin to regulate the difference in loss between preferred and dispreferred responses, as detailed below:

\begin{equation}
    \begin{split}
        & \mathcal{L}_{cal}(\pi_\theta)=\mathbb{E}_{x,y_w,y_l\sim D} \\
        & [\max(0,\beta-\log\pi_\theta (y_w|x)+\log \pi_\theta (y_l|x)]
    \end{split}
\end{equation}


\noindent where $\beta$denotes the margin ensuring that the log probability of the preferred response surpass that of the dispreferred response by at least $\beta$. Furthermore, SLiC includes a cross-entropy component for responses generated by the reference model, with the goal of minimizing substantial divergence from the reference model, as outlined below:

\begin{equation}
\begin{split}
    & \mathcal{L}_{SLiC}(\pi_\theta,\pi_{ref})=\mathcal{L}_{cal}(\pi_\theta)+\lambda\mathbb{E}_{x\sim D,y\sim\pi_{ref}(x)} \\
    & [-\log \pi_\theta (y|x)]
\end{split}
\end{equation}

\subsection{Kahneman-Tversky Optimization}


The Kahneman-Tversky Optimization (KTO) method posits that pairs of preferences might be unnecessary and advocates for the direct maximization of utility derived from LLMs outputs, rather than focusing on maximizing the log-likelihood of preferences, as described below:

\begin{equation}
    \mathcal{L}_{KTO}=\mathbb{E}_{(x,y)\sim D}[w(y)(1-\hat{h}(x,y;\beta ) )]
\end{equation}

\noindent where $h(x,y;\beta )$ indicates a human value function, which can be expressed as follows:

\begin{equation}
    h(x,y;\beta)=\begin{cases}
 \sigma (g(x,y;\beta)) & \text{ if } y\sim y_{w}|x \\
 \sigma (-g(x,y;\beta)) & \text{ if } y\sim y_{l}|x
\end{cases}
\end{equation}

\noindent where $\sigma$ is a logistic function, and $g(x,y;\beta)$ can be defined as follows:

\begin{equation}
\begin{split}
    & g(x,y;\beta)=\beta\log\frac{\pi_\theta(y|x) }{\pi_{ref}(y|x)} -\mathbb{E}_{x'\sim D}\\
    & [\beta KL(\pi_\theta ||\pi_{ref})]
\end{split}
\end{equation}


\noindent where $KL(\cdot)$ represents the Kullback-Leibler divergence function used to limit the deviation of the LLM from the reference model, and $w(y)$ within the loss function $\mathcal{L}_{KTO}$ is specified as follows:

\begin{equation}
    w(y)=\begin{cases}
 \lambda_D & \text{ if } y\sim y_{w}|x \\
 \lambda_U & \text{ if } y\sim y_{l}|x
\end{cases}
\end{equation}

\noindent where $\lambda_D$ and $\lambda_U$ are loss weights.

\section{GPT-4 prompts for Evaluating}
\label{appendix:prompts}

In our work, GPT-4 serves as a surrogate for assessing the responses from LLMs in comparison to the baselines from the Antropic-HH dialogue and TL;DR summarization datasets. This section outlines the prompts utilized to derive win rates for our experimental analysis.

\subsection{Prompt for Summarization}
Following \cite{DBLP:conf/nips/RafailovSMMEF23}, we use the following prompt to evaluate the answers of LLMs:

\textit{Which of the following summaries does a better job of summarizing the most import points in the given forum post, without including unimportant or irrelevant details? A good summary is both precise and concise.}

\textit{Post:}
\textit{<post>}

\textit{Summary A:}
\textit{<Summary A>}

\textit{Summary B:}
\textit{<Summary B>}

\textit{FIRST provides a one-sentence comparison of the two summaries, explaining which you prefer and why. SECOND, on a new line, state only "A" or "B" to indicate your choice. Your response should use the format:}
\textit{Comparison: <one-sentence comparison and explanation>}
\textit{Preferred: <"A" or "B">}

\subsection{Prompt for Dialogue}

\textit{Here is a conversation between Human and Assistant.}

\textit{Conversation: <Conversation>}

\textit{Assistant can now choose from different responses.}

\textit{Response 1: <Response 1>}

\textit{Response 2: <Response 2>}

\textit{Between these two responses, which response is more helpful, harmless, and concise? Please only answer RESPONSE 1 or RESPONSE 2.}

\subsection{Impact of Data Volume and Update Steps}
To address the potential impact of preference data differences and update steps on performance, we conducted additional experiments. We explored the effects of varying data volumes and update steps on model performance using LLaMA-7B on the TLDR dataset.


\begin{table}[h]
    \centering
    \resizebox{0.8\linewidth}{!}{
    \begin{tabular}{@{}c|cc@{}}
    \toprule
    \multirow{2}{*}{\textbf{Method}}  & \multicolumn{2}{c}{TL;DR summarization dataset}  \\
    & FULL & HALF  \\
    \midrule
     DPO   & 81.0\%  & 78.8\%   \\
     DPO + SPO   & 83.6\% & 81.4\%  \\
    \bottomrule
    \end{tabular}
    }
    \caption{Performance comparison of DPO and DPO+SPO methods with varying data sizes. "Full" refers to the complete dataset, while "Half" indicates using half of the dataset. }
    \label{table:data volume}
\end{table}



\begin{table}[h]
    \centering
    \resizebox{0.8\linewidth}{!}{
    \begin{tabular}{@{}c|cc@{}}
    \toprule
    \multirow{2}{*}{\textbf{Method}}  & \multicolumn{2}{c}{TL;DR summarization dataset}  \\
    & 1 EPOCH & 2 EPOCH  \\
    \midrule
     DPO   & 81.0\%  & 80.0\%   \\
     DPO + SPO   & 83.6\% & 81.6\%  \\
    \bottomrule
    \end{tabular}
    }
    \caption{Performance comparison of DPO and DPO+SPO methods with different update steps. "1 EPOCH" and "2 EPOCH" denote the number of training iterations. }
    \label{table:update step}
\end{table}

From the Table~\ref{table:data volume}, we can see that under the setting of half the data volume, both DPO and DPO+SPO methods show a decline, but DPO+SPO still maintains better performance than DPO. From the Table~\ref{table:update step}, we can see that in different update steps settings, too large update steps lead to overfitting in DPO, but DPO+SPO still performs better than DPO. Overall, under different data volumes and update steps settings, the trend of DPO+SPO is consistent with DPO, indicating that data volume and update steps have little impact on our method. 
\subsection{Impact of Different Module of Self-Supervised Classcification Module}
In our self-supervised training, we utilized a classification model with a two-layer MLP and positional encoding. This design is based on two hypotheses:
\begin{itemize}
\item Compared to directly inputting embeddings into the classification head, using a two-layer MLP helps mitigate the negative impact of self-supervised loss on the embedding distribution, thereby improving the effectiveness of the self-supervised embeddings.
\item We hypothesized that the method of keyword deletion might lead to semantic discontinuity, causing the model to struggle with learning preferences effectively. Therefore, we added the original positional encoding to the latent embeddings, hoping that the model could better learn preferences.
\end{itemize}

\begin{table}[h]
    \footnotesize
    \centering
    \begin{tabular}{c|ccc}
    \toprule
     Method & Classifier & PE & WR  \\
    \midrule
     DPO   & -  & - & 81.0 \\
     \midrule
     \multirow{4}{*}{DPO + SPO}   & a FC layer & \XSolidBrush & 81.5\\
            & a FC layer & \Checkmark & 82.7\\
            & a two-layer MLP + a FC layer & \XSolidBrush & 83.1\\
            & a two-layer MLP + a FC layer & \Checkmark & 83.6\\

    \bottomrule
    \end{tabular}
    \caption{Comparison of Win Rates for Different Classifier Configurations and Positional Encoding in Self-Supervised Training on Llama. FC indicates Fully-Connected, PE means Position Encoding, WR stands for Win Rate (\%).}
    \label{table:self supervised module}
\end{table}

Based on these two hypotheses, we conducted experiments on LLaMA-7B on TLDR dataset. From the Table~\ref{table:self supervised module}, we can see that using only the FC layer resulted in a 0.5\% improvement in SPO, which, although validating the method's effectiveness, is quite trivial. Using the FC layer with positional encoding resulted in a 1.7\% improvement in SPO, indicating that positional encoding in the latent embeddings could help the model better understand preferences, thereby enhancing performance. When we added an additional projection layer, i.e., a two-layer MLP before the FC layer (without positional encoding), we observed a 2.1\% improvement, which is a 1.6\% increase over using the FC layer alone, demonstrating the effectiveness of the two-layer MLP. Finally, when we combined the two-layer MLP with positional encoding, we observed a maximum improvement of 2.6\%. This experiment demonstrates the effectiveness of our designed classification module.

\section{Cases of different extracting method}
\label{app:case_extracting_method}

We employ various extraction techniques ((i.e., YAKE \cite{campos2020yake}, RAKE \cite{rose2010automatic} and PositionRANK \cite{florescu2017positionrank})) to identify key content on HH dataset, with illustrative examples provided below.

\begin{itemize}
    \item \textit{\textbf{Raw response}: "I'm sorry, this doesn't seem like the kind of thing I'm built to handle. Can you explain to me more what you mean? Is it really that loud?"}

    \item \textit{\textbf{RAKE}: Key content: ["Can you explain to me more what you mean", "doesn't seem like the kind of thing I'm built", "Is it really that loud"]}

    \item \textit{\textbf{YAKE}: Key content: ["kind of thing I built to handle", "built to handle", "kind of thing"]}

    \item \textit{\textbf{PositionRank}: Key content: ["kind", "thing", "handle"]}
\end{itemize}

Based on the above samples, we can see that RAKE tends to extract more continuous key content while YAKE and PositionRANK generate sparse key contents. 


\end{document}